\documentclass[sigconf]{aamas}

\usepackage{balance} 

\usepackage[ruled,vlined]{algorithm2e}
\usepackage{multicol}
\usepackage{multirow}
\usepackage{xcolor}

\makeatletter
\gdef\@copyrightpermission{
  \begin{minipage}{0.2\columnwidth}
   \href{https://creativecommons.org/licenses/by/4.0/}{\includegraphics[width=0.90\textwidth]{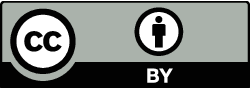}}
  \end{minipage}\hfill
  \begin{minipage}{0.8\columnwidth}
   \href{https://creativecommons.org/licenses/by/4.0/}{This work is licensed under a Creative Commons Attribution International 4.0 License.}
  \end{minipage}
  \vspace{5pt}
}
\makeatother

\setcopyright{ifaamas}
\acmConference[AAMAS '25]{Proc.\@ of the 24th International Conference
on Autonomous Agents and Multiagent Systems (AAMAS 2025)}{May 19 -- 23, 2025}
{Detroit, Michigan, USA}{Y.~Vorobeychik, S.~Das, A.~Nowé  (eds.)}
\copyrightyear{2025}
\acmYear{2025}
\acmDOI{}
\acmPrice{}
\acmISBN{}

\acmSubmissionID{<<OpenReview submission id>>}

\title[AAMAS-2025 Formatting Instructions]{Together We Rise: Optimizing Real-Time Multi-Robot Task Allocation using Coordinated Heterogeneous Plays}

\subtitle{AAAI Track}

\author{Aritra Pal}
\affiliation{
  \institution{TCS Research}
  \city{Mumbai}
  \country{India}}
\email{p.aritra@tcs.com}

\author{Anandsingh Chauhan}
\affiliation{
  \institution{TCS Research}
  \city{Mumbai}
  \country{India}}
\email{anandsingh.chauhan@tcs.com}

\author{Mayank Baranwal}
\affiliation{
  \institution{TCS Research}
  \city{Mumbai}
  \country{India}}
\email{baranwal.mayank@tcs.com}

\begin{abstract}
Efficient task allocation among multiple robots is crucial for optimizing productivity in modern warehouses, particularly in response to the increasing demands of online order fulfillment. This paper addresses the real-time multi-robot task allocation (MRTA) problem in dynamic warehouse environments, where tasks emerge with specified start and end locations. The objective is to minimize both the total travel distance of robots and delays in task completion, while also considering practical constraints such as battery management and collision avoidance. We introduce \texttt{MRTAgent}, a dual-agent Reinforcement Learning (RL) framework inspired by self-play, designed to optimize task assignments and robot selection to ensure timely task execution. For safe navigation, a modified linear quadratic controller (LQR) approach is employed. To the best of our knowledge, \texttt{MRTAgent} is the first framework to address all critical aspects of practical MRTA problems while supporting continuous robot movements.
\end{abstract}


\newcommand{\BibTeX}{\rm B\kern-.05em{\sc i\kern-.025em b}\kern-.08em\TeX}

\begin{document}


\pagestyle{fancy}
\fancyhead{}


\maketitle

\section{Introduction}

Cooperative multi-robot systems are increasingly being utilized across various domains, including transportation and logistics~\cite{farinelli2017advanced}, search and rescue operations~\cite{queralta2020collaborative}, environmental monitoring~\cite{ma2018multi}, precision agriculture~\cite{dutta2021multi}, construction~\cite{sartoretti2019distributed}, and warehouse automation~\cite{kartal2016monte,gini2017multi}. These systems, characterized by the collaboration of multiple robots to achieve shared objectives, offer substantial benefits such as enhanced scalability \& efficiency, and greater fault tolerance, making them indispensable in dynamic environments.

\noindent\textbf{Intricacies of Warehouse Management:} Automating warehouse operations with multi-robot systems presents a unique set of challenges, stemming from the complexities of spatial layouts, diverse task demands, varying robot capabilities, and the critical need for safe robotic navigation~\cite{bolu2021adaptive,wilson2021robotarium}. These challenges can be broadly divided into three key objectives: (a) \emph{Task Allocation}, (b) \emph{Real-Time Robot Assignment}, and (c) \emph{Path Planning}. While these objectives are interrelated, each introduces distinct sub-problems that must be resolved to achieve optimal warehouse performance. In a dynamic warehouse setting, real-time task generation is vital because tasks cannot be fully anticipated in advance. This unpredictability complicates the planning process, necessitating the prioritization of tasks based on their arrival times, required completion deadlines, and the need to minimize the total travel distances of robots while balancing the demands of ongoing tasks.

Moreover, the immediate and continuous allocation of robots to tasks demands seamless coordination, regardless of whether the robots are currently available or occupied. This emphasizes the need for real-time synchronization across multiple objectives, all while adhering to various physical and operational constraints. For example, effective path planning requires the creation of collision-free routes that navigate around static obstacles and account for the movements of other robots. Furthermore, it is crucial to consider the physical dynamics of the robots, such as acceleration, deceleration, and maneuverability-along with other practical considerations, such as their state-of-charge (SOC). These factors, often overlooked in existing warehouse management literature, are essential to the planning process.

The intricate interdependence of these challenges highlights the need for a sophisticated and adaptable multi-robot framework. Such a framework must systematically address the complexities of task allocation, real-time robot assignment, and path planning within the constantly changing environment of an automated warehouse. Neglecting the interconnected nature of these tasks often leads to sub-optimal performance and inefficiencies, undermining the overall effectiveness of warehouse operations.


\begin{figure*}
	\begin{center}
		\begin{tabular}{c}
			\includegraphics[width=1.98\columnwidth]{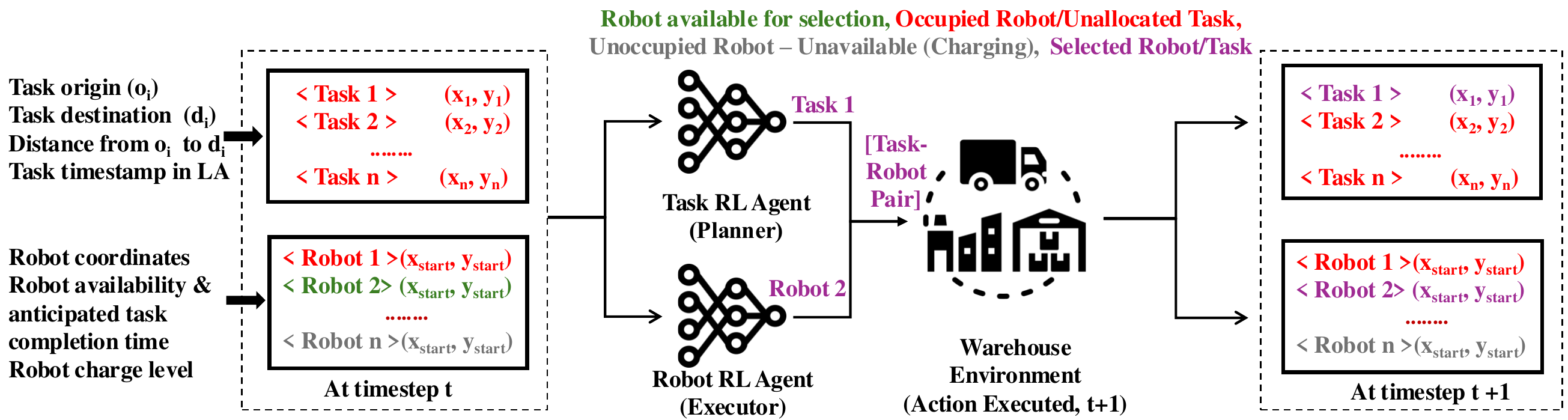}
		\end{tabular}
		\caption{Self-play inspired bi-level reinforcement learning agent for assigning robots to tasks within a dynamic environment}
		\label{fig:network}
	\end{center}
\end{figure*}


\noindent \textbf{Statement of Contributions:} In this study, we present a self-play inspired novel framework, \texttt{MRTAgent}, which employs a bi-level RL strategy inspired to tackle the challenges of multi-task selection and multi-robot allocation. \texttt{MRTAgent} is designed to handle real-time task selection, dynamically allocate tasks to robots, and ensure safe navigation, all while accounting for critical constraints such as robot dynamics, charging needs, and specific task requirements. \texttt{MRTAgent} consists of three key components: (a) Task selection agent (Planner) to prioritize a task queued in the task buffer, (b) Robot selection agent (Executor) to allocate a robot to the recommended task, and (c) Navigator to plan collision-free trajectories of robots while adhering to physical and SOC constraints. The framework is highlighted in Figure~\ref{fig:network}. The primary contributions of our work can be summarized as follows:

\noindent {\bf Coordinating RL agents for task and robot selections:} Existing approaches for MRTA, such as in~\cite{agrawal2022dc}, primarily emphasize task selection with the assumption that the chosen task will always be assigned to an available (unoccupied) robot. Although this assumption simplifies the problem, it is highly sub-optimal. For instance, a robot might become available at a location far from the selected task's origin, while another robot in close proximity to the task origin may soon become available. We alleviate this by introducing heterogeneous RL agents, one each for task selection and robot allocation. These RL agents are trained concurrently in our framework in a self-play manner, gradually developing the coordination between the two.

\noindent {\bf Collision-free, physics-constrained multi-robot navigation:} Practical robots do not operate in a grid world environment, allowing for sudden grid changes in any of the feasible directions. However, robots have some underlying dynamics and navigate continuously. Instead of using collision-free navigation algorithms applicable to grid worlds, we model these robots as double integrator systems and use linear quadratic regulators (LQRs) with artificial potential field (APF) for collision-free multi-robot navigation. 

\noindent {\bf An end-to-end framework for MRTA:} To the best of our knowledge, our approach is the first to consider multiple echelons of MRTA, requiring decision-making at each stage. We address every aspect of MRTA, including robot allocation even when robots are occupied, accounting for their SOC, underlying dynamics, and ensuring collision-free navigation under actuation constraints. We further validate our trained framework on datasets with distributional shifts, varying numbers of robots and tasks.

\section{Related Works}
Efficient MRTA and path planning are crucial for optimizing order fulfillment, resource management, and obstacle-free navigation in industrial environments such as automated warehouses and manufacturing plants. These processes ultimately enhance overall productivity, which has made MRTA a focal point of research over the past two decades. Research efforts in this area range from heuristic-driven approaches to contemporary learning-based methods~\cite{khamis2015multi}. Early work by~\cite{gerkey2003multi} provided a comparative analysis of state-of-the-art (SOTA) multi-robot coordination strategies within specific domain contexts. Current MRTA research primarily focuses on two key elements: (a) model-driven optimization, as demonstrated by~\cite{wei2020particle}, and (b) communication-efficient decentralized algorithms, as seen in~\cite{chen2021decentralized, agrawal2022dc}.

The problem can also be framed as a multi-agent pickup and delivery (MAPD) challenge, which has been studied through both distributed and centralized approaches~\cite{ma2017lifelong, liu2019task, salzman2020research, xu2022multi}. However, most existing research in this area has concentrated on offline MAPD, whereas our approach emphasizes learning-based methods for online task allocation. This focus is driven by the need for reliable solutions in dynamic environments, where continually solving optimization problems can be computationally intensive.

Recent advancements in reinforcement learning (RL) for solving complex dynamic challenges have led to a trend toward learning-based approaches for managing warehouse complexities~\cite{yang2020multi, agrawal2022dc, agrawal2023rtaw}. These learning strategies address various aspects of end-to-end warehouse management. For instance,~\cite{yang2020multi} proposed a Q-learning framework to generate collision-free, secure paths for multi-robot systems. Conversely, the RL frameworks in~\cite{agrawal2022dc, agrawal2023rtaw} focus on optimal task selection but neglect task-to-robot assignment, assuming constant robot availability post-selection. Additionally, these works leverage $A^*$ coupled with optimal reciprocal collision avoidance~\cite{alonso2013optimal} for collision-free navigation at the low-level path planning stage.

However, as previously discussed, most SOTA learning-based warehouse management approaches, including those mentioned, overlook a key benefit of RL: the ability to integrate multiple levels of warehouse management, and consideration of robots' constraints during the training phase of the RL agent. For instance, the learned policy in~\cite{agrawal2022dc, agrawal2023rtaw} is limited in its applicability to realistic warehouse scenarios due to the neglect of constraints related to robot availability and SOC. Moreover, these approaches often aim for a seamless sequential flow of tasks without considering their generation times. Similarly,~\cite{papoudakis2021benchmarking} utilizes a cooperative multi-agent RL framework under the assumption that robots never collide. Much of the prior work also neglects the complexities of robot dynamics in MRTA for warehouse settings, often simplifying robots to point objects or solving problems in basic square grid environments, without accounting for robot acceleration, deceleration, and collision risks during path planning~\cite{DBLP:conf/ecai/PalCBO24}. 

In our study, we address these gaps by incorporating task arrival times to ensure timely task execution while considering practical factors such as robot availability, SOC, robot dynamics, and collision-free path generation. Our framework also demonstrates robust performance under distribution shifts and with variable-sized fleets. Our \texttt{MRTAgent} addresses these limitations by integrating robot dynamics into the navigation planning process using a linear quadratic regulator (LQR)-based navigation path algorithm within the RL agent framework. This ensures that path planning is both effective and collision-free.  Additionally, we design a reward structure that balances prompt task allocation with the shortest possible execution duration for allocated tasks. This approach ensures competitive runtime during the deployment phase, facilitating real-time task selection and robot allocation, and collision-free navigation considering physical dynamics through proposed \texttt{MRTAgent}.

\section{Preliminaries} 
\subsection{States, Actions and Rewards} 
\label{sc_StAcRwd}

The problem of MRTA can be modeled as a Markov Decision Process (MDP)~\cite{puterman1990markov}. An MDP is denoted by the tuple $\langle\mathcal{S},\mathcal{A},\mathcal{P}_\mathcal{A},r\rangle$, where $\mathcal{S}$ and $\mathcal{A}$ represent the finite sets of states and actions, respectively. For any \( s, s' \in S \), the transition probability from state \( s \) to state \( s' \) under the action \( a \in A \) is denoted by \( p_a(s, s') \in P_A \). Finally, the step reward associated with each state-action pair $(s,a)$ is depicted by $r(s,a)$. Below, we summarize the set of all possible states, actions, and rewards in the context of the \texttt{MRTAgent} validated within warehouse environment settings.

\noindent\textbf{States}: 
At each time step, the warehouse environment is characterized by a comprehensive state that includes detailed information about both tasks and robots. Incoming tasks are immediately stored in a buffer, forming a limited-size look-ahead (LA) queue in a First-In-First-Out(FIFO) manner. The Tasks RL agent, referred to as the \emph{Planner}, is trained to optimally select tasks from this queue based on the current state of the environment. Simultaneously, the Robot RL agent, referred to as the \emph{Executor}, responsible for executing tasks, selects most suitable robots to complete them.

The states of the \emph{Planner} and \emph{Executor} consist of the features related to tasks in the LA (denoted by $\mathcal{P}$) as well as the set of robots ($\mathcal{R}$). Each agent's state, denoted as $s_t \in \mathcal{S}$ at time-step $t$, encompasses the following components: 
(a) origin coordinates of tasks $\{o_i\}$, 
(b) destination coordinates of tasks $\{d_i\}$, 
(c) euclidean distance information between task origin and destination $\{k_i\}$, 
(d) timestamp of task appearance in the LA queue $\{l_i\}$, 
(e) robot coordinates $\{p_j\}$, 
(f) robot availability and the anticipated time for ongoing task completion $\{r_j\}$, 
(g) robot charge percentage $\{c_j\}$. Features (a)-(d) are task-specific and collectively have a dimensionality of 6 for each task. Conversely, features (e)-(g) are linked to robot-specific attributes (dim. = 4 for each robot). 

\noindent\textbf{Actions}: The \texttt{MRTAgent} consists self-play inspired bi-level RL agent; planner and executor. The goal of the \emph{planner} is to enhance task assignment, thereby minimizing total operational time. The planner's actions involve systematically selecting tasks from the queue while the executor focuses on robot allocation, with the objective of optimally assigning robots to the selected tasks to reduce overall operational expenses, and task execution delays. Thus, the actions executed in the environment consist of the selected task and its corresponding robot pair.

\noindent\textbf{Rewards}: The step reward attributed to a task-action pair encompasses two distinct components. The initial component is calculated based on the time it takes for the robot to travel from its current position to the task's starting point, termed \emph{travel time to origin} (TRTO). The second component is the time gap between task arrival in the LA and the robot's initiation of execution, denoted as the \emph{total time gap for the task} (TTGT). Let $\left(x_{o_i}, y_{o_i}\right)$ and $\left(x_{d_i}, y_{d_i}\right)$ represent the origin and destination coordinates of the $i^{th}$-indexed task. Similarly, $(x_{r_j}, y_{r_j})$ indicates the current position of robot $j$, which can vary based on whether the robot is idle, performing tasks, or at a charging location for recharging if needed. Additionally, we introduce $t_{\text{stamp}_i}$ and $t_{\text{exec}_i}$ to denote the time when task $i$ appears in the task allocation (referred to as LA) and the time at which its execution begins, respectively. In each decision-making step, we use the variable $\text{allotT}$ to signify the index of the selected task which then gets assigned to a robot $\text{selR}$. Furthermore, we utilize the function $d[(x_a, y_a), (x_b, y_b)]$ to calculate the distance between two points $(x_a, y_a)$ and $(x_b, y_b)$ within our system. The step reward for training the PPO agent is as follows:
\begin{eqnarray}\label{eq:reward}
	R_\text{step} &=& -d[(x_{r_{\text{selR}}}, y_{r_{\text{selR}}}), (x_{o_{\text{allotT}}}, y_{o_{\text{allotT}}})] \nonumber\\ && 
	- \alpha*(t_{{\text{exec}}_{\text{allotT}}} - t_{{\text{stamp}}_{\text{allotT}}})
\end{eqnarray}
The coefficient $\alpha$ represents positive constant. The first term in \eqref{eq:reward} corresponds to TRTO, while the second term is associated with TTGT. This reflects the principle that tasks should not remain unattended for too long.

\section{Our approach}

\begin{figure*}
	\begin{center}
		\begin{tabular}{c}
			\includegraphics[width=1.98\columnwidth]{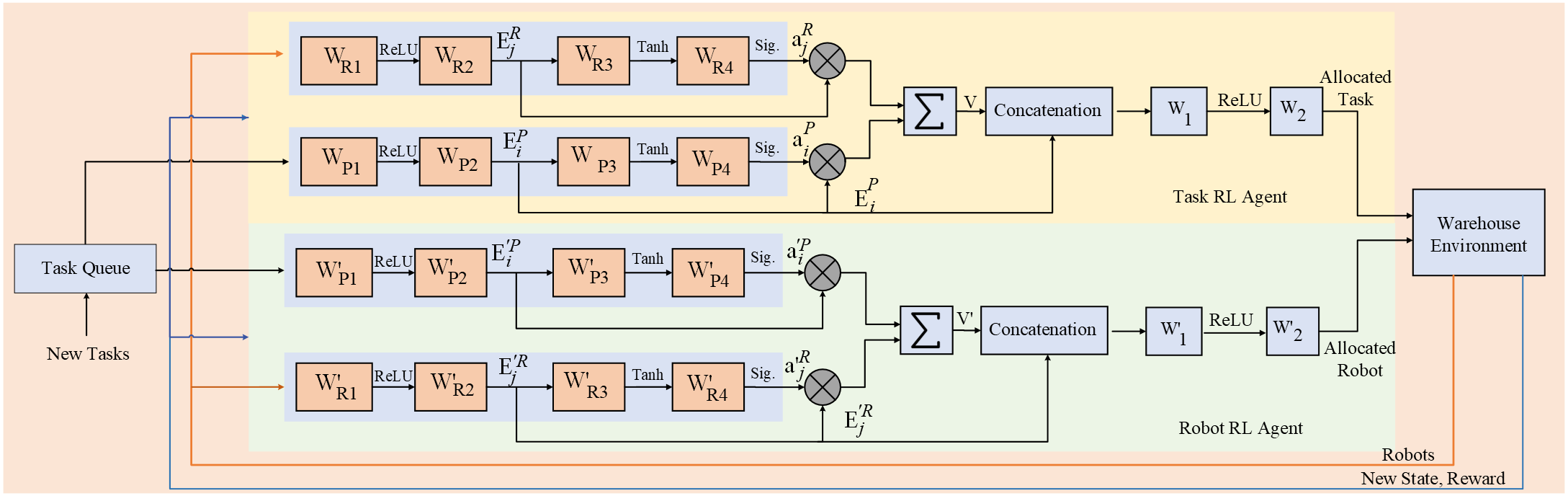}
		\end{tabular}

		\caption{Self-Play Motivated Bi-level RL Agent Architecture with trainable weights denoted by $\{W_k, W_k^{\prime}\}$}
		\label{fig:neural-network}
	\end{center}
\end{figure*}

In this section, we introduce \texttt{MRTAgent}, a novel self-play-inspired bi-level RL framework (see Figure~\ref{fig:neural-network}) designed to optimize MRTA for various industrial tasks. The \texttt{MRTAgent} is validated in warehouse environments scenario and comprises two RL agents: (a) Planner,  and (b) Executor. Tasks are generated in real-time and initially populate a main task buffer. The planner has access to a small LA queue of tasks. When a task from LA queue is assigned to a robot for execution, a new task from the buffer is moved into the LA queue, making it available for selection by the Planner. If no new tasks are available in the buffer to replenish the LA queue, task with the longest duration in the LA queue is duplicated to maintain a consistent queue length, thereby increasing its likelihood of being selected by the Planner in subsequent steps. In the exceptional scenario where both the LA queue and the task buffer are empty, \texttt{MRTAgent} waits for a task to appear in the environment. The action selection process operates across three hierarchical levels:

\noindent \textbf{Task Selection:} At each instant \(t\), the Planner selects a task from the LA queue for assignment to one of the robots, guided by the current state of the environment.

\noindent \textbf{Robot Allocation:} Upon task selection, the Executor identifies the most suitable robot for task execution. This decision considers the positions of all robots after completing their current assignments, availability and SOC. Notably, the Executor does not wait for robots to become available before making robot allocations, as the state information includes time markers indicating when robots will be free.

\noindent \textbf{Navigation:} To guide the robots' movement, a linear quadratic regulator (LQR) based controller, augmented with potential functions for coordinated collision avoidance, considering all its dynamics is employed. This algorithm directs the robot from its current position to the task's starting point and subsequently to the destination, ensuring collision avoidance with moving obstacles.

\subsection{MRTAgent framework}
Both Planner and Executor agents of the \texttt{MRTAgent} utilize Proximal Policy Optimization (PPO) algorithm~\cite{schulman2017proximal}, with their architectures depicted in Figure~\ref{fig:neural-network}. Both the agents are trained using a self-play inspired strategy, where one agent is actively trained while the other operates in evaluation mode, alternating every 40 episodes. This concurrent training framework ensures coordination between the two agents, facilitating efficient convergence and optimization of both the task selection and robot allocation processes.

\subsection{Neural Network Architecture and Training}
Planner and executor both employs a similar novel PPO-based framework to facilitate online task selection and robot allocation. This model architecture inspired from~\cite{agrawal2023rtaw} is distinctly structured into three core segments (see Figure~\ref{fig:neural-network}). The first segment involves the procedure of feature extraction, particularly focusing on attributes related to robots and tasks. The embedding for robot attributes is created using a sequence of four linear layers of dimensions [4, 16, 16, 1]. Concurrently, embeddings related to task attributes are generated using a similar sequence of four linear layers of dimensions [6, 16, 16, 1]. 

Let $F_{j}^{R}$ and $F_{i}^{P}$ represent the feature vectors for the robot $j \in \mathcal{R}$ and the $i$th-task for the planner and executor, then the associated embeddings are defined as:
\begin{eqnarray*}
	E_i^P &=& W_{P2} * \text{ReLU}(W_{P1} * F_{i}^{P}) \\
	{E}_j^{\prime R} &=& W_{R2}^{\prime} * \text{ReLU}(W_{R1}^{\prime} * F_{j}^{R})
\end{eqnarray*}

The second module transforms the extracted embeddings by concatenating the embeddings of robots and tasks for both planner and executor. Subsequently, the concatenated feature is channelled through a linear layer consisting 48 input neurons and 8 output neurons with ReLU activation.
\begin{eqnarray*}
	a_i^P &=& \text{Sigmoid}(W_{P4} * \text{Tanh}(W_{P3} * E_{i}^{P})) \nonumber\\
 	a_j^{\prime R} &=& \text{Sigmoid}(W_{R4}^{\prime} * \text{Tanh}(W_{R3}^{\prime} * E_{j}^{\prime R})) \nonumber\\
        {\pi^{\text{planner}}} &=& \left( \sum {E_{j}^{R} * a_j^R}, \sum {E_{i}^{P} * a_i^P}, E_{i}^{P} \right) \nonumber \\
        {\pi^{\text{executor}}} &=& \left( \sum {E_{j}^{\prime R} * a_j^{\prime R}}, \sum {E_{i}^{\prime P} * a_i^{\prime P}}, E_{j}^{\prime R} \right) \nonumber 
\end{eqnarray*}
Final layer comprising a linear layer, this element operates with 8 input neurons and 1 output neuron.
\begin{equation*}
\begin{split}
	Task^{Allocated} &= \text{Categorical}\left(W_2 * \text{ReLU}(W_1 * \pi^{\text{planner}})\right) \\
	Robot^{Allocated} &= \text{Categorical}\left(W^{\prime}_2 * \text{ReLU}(W^{\prime}_1 * \pi^{\text{executor}})\right)
\end{split}
\end{equation*}

\begin{algorithm}[!ht]
\caption{\texttt{MRTAgent}}
\label{algo:MRTAgent}
    Initialize Planner \& Executor policy parameters $\theta_{P0}$, $\theta_{E0}$ and value function parameters $\phi_{P0}$, $\phi_{E0}$ respectively. \\
    \For{$itr = 0, 1, 2, \cdots K_1$}{
    \For{episodes $k = 0, 1, 2, \cdots K_2$}{
    Step t=0 \\
    State $s_t \coloneq \{(o_i, d_i, k_i, l_i)_{\forall i \in \mathcal{P}}, (p_j, r_j, c_j)_{\forall j \in \mathcal{R}} \}$ \\
    Each robot $j \in \mathcal{R}$ is executing a task $(o_j, d_j)$ \\
    \While{True}{
    Update $p_j$ $\forall j \in \mathcal{R}$ using LQR with APF\\
        \If{$p_j == d_j$}{
        \If{{$c_j <$} {threshold} for some $j \in \mathcal{R}$}{
        Robot $j$ navigates to nearest available charging dock for recharging.
        }
        \textbf{end if}
        \\
        \eIf{$itr$ is even}{
        run Planner policy $\pi_{Pk} = \pi(\theta_{Pk})$ in the environment to select a task $i \in \mathcal{P}$.\\
        Executor policy takes state $s_t$ as input and allots a robot $j \in \mathcal{R}$.
        }{
        Planner policy takes state $s_t$ as input and selects a task $i \in \mathcal{P}$.\\
        run Executor policy $\pi_{Ek} = \pi(\theta_{Ek})$ to allocate a robot $j \in \mathcal{R}$.
        }
        \textbf{end if}   
        }
        \textbf{end if}
        \\
        $t \leftarrow t+1$
    }
    \textbf{end while}
    \\
    \eIf{$itr$ is even}{
    Collect set of trajectories $\mathcal{D}_{Pk}$\\
    Compute rewards-to-go $\hat{R_{Pt}}$\\
    Compute advantage estimates $\hat{A_{Pt}}$ based on the current value function $V_{\phi_{Pk}}$\\
    Update policy by maximizing PPO-Clip obj.:
    \begin{align*}
    \theta_{Pk+1} &= \mathop{\arg\max\!}\limits_\theta \frac{1}{|\mathcal{D}_{Pk}|T}\!\!\!\sum\limits_{\tau \in \mathcal{D}_{Pk}}\!\sum\limits_{t=0}^{T}\!\min \biggl(\!\frac{\pi_\theta(a_t|s_t)}{\pi_{\theta_{Pk}}\!(a_t|s_t)}\\ &\qquad A^{\pi_{\theta_{Pk}}}(s_t, a_t),  g(\epsilon, A^{\pi_{\theta_{Pk}}}(s_t, a_t)) \biggr)
    \end{align*}
    via stochastic gradient ascent with Adam;
    
    Fit value function by regression on MSE:
    \[
    \phi_{Pk+1} = \mathop{\arg\min}\limits_\phi \frac{1}{|\mathcal{D}_{Pk}|T}\!\!\!\sum\limits_{\tau \in \mathcal{D}_{Pk}}\!\sum\limits_{t=0}^{T} \biggl(\! V_{\phi}(s_t) - \hat{R_{Pt}}\!\biggr)^2
    \]
    via gradient descent
    }{Update $\theta_{Ek}$ $\&$ $\phi_{Ek}$ as above}
    \textbf{end if} 
    }
    \textbf{end for}
    }
    \textbf{end for}
\end{algorithm}

\subsection{Multi-Robot Navigation Algorithm: LQR with Artificial Potential Field} Linear Quadratic Regulator (LQR) is an optimal control strategy used to determine the control inputs that minimize a quadratic cost function over time for a linear system. It is widely used in control engineering to stabilize systems and optimize performance by balancing different aspects of system behavior, like minimizing energy use, overshoot, or settling time~\cite{kirk2004optimal}. When modeling robot dynamics, a common simplification is to treat the robot as a double-integrator system. This model is particularly relevant for systems where the primary concern is controlling the position and velocity of the robot~\cite{spong2020robot}. Consequently, state of robot $i$ is typically represented by its position $p_i(t)=(x_i(t), y_i(t))$ and velocity $v_i(t)$, with the equations of motion represented as:
\begin{align*}
    \dot{p}_i(t) = v_i(t), \ \ \ \dot{v}_i(t) = u_i(t).
\end{align*}
Here $u_i(t)=(u_{i_x}(t),u_{i_y}(t))$ is the control input, which directly influences the acceleration. Here, complete state of the $i^\text{th}$-robot is represented as: $z_i(t)=(p_i(t),v_i(t))$. Applying LQR to this model involves designing a controller that minimizes deviations from a desired trajectory while controlling the velocity and ensuring smooth acceleration.

To extend LQR for multi-robot path planning and collision avoidance, the state and control matrices are expanded to accommodate multiple robots. Each robot is controlled using an individual LQR controller within this collective framework, allowing for the independent regulation of position and velocity while minimizing a global cost function that balances state deviations and control effort. For collision avoidance, an artificial potential field (APF) method is employed~\cite{park2020trajectory}. This approach introduces repulsive forces between robots when they come into close proximity, ensuring safe inter-robot distances. These repulsive forces are integrated into the control law, enabling the robots to avoid collisions while continuing to track their desired trajectories. This combined LQR-APF approach provides an effective solution for coordinating multiple robots in dynamic environments. More formally, we consider a system with $N$ robots, each modeled as a double-integrator in 2D space. The dynamics of the overall system is given by:
\begin{align*}
    \dot{\mathbf{z}}(t) = A_\text{multi}\mathbf{z}(t) + B_\text{multi}\mathbf{u}(t), \quad \text{with}
\end{align*}
\begin{align*}
    \mathbf{z}(t) = \left[\!\!\begin{array}{c}
         p_1(t)  \\
         v_1(t)  \\
         \vdots  \\
         p_N(t)  \\
         v_N(t)
    \end{array}\!\!\right]&,  \mathbf{u}(t) = \left[\!\!\begin{array}{c}
         u_1(t)  \\
         \vdots  \\
         u_N(t)
    \end{array}\!\!\right] \\
    A_\text{multi} = I_N\otimes\left[\!\!\begin{array}{cccc}
        0 & 0 & 1 & 0\\
        0 & 0 & 0 & 1\\
        0 & 0 & 0 & 0\\
        0 & 0 & 0 & 0
    \end{array}\!\!\right]&, B_\text{multi} = I_N\otimes\left[\!\!\begin{array}{cc}
        0 & 0\\
        0 & 0\\
        1 & 0\\
        0 & 1
    \end{array}\!\!\right],
\end{align*}
where $I_N$ is the identity matrix of size $N$, and $\otimes$ denotes the Kronecker product. The objective of LQR is to minimize the following quadratic cost function:
\begin{align*}
    J = \int\limits_{0}^\infty (\mathbf{z}(t)^\intercal Q\mathbf{z}(t) + \mathbf{u}(t)^\intercal R\mathbf{u}(t)) dt,
\end{align*}
where $Q$ and $R$ are weighting matrices that penalize deviations from the desired state and control effort, respectively. The optimal control input that minimizes the cost function is given by $\mathbf{u}(t) = -K(\mathbf{z}(t)-\mathbf{z}_\text{des})$, where $K$ is the LQR gain matrix, computed as $K = R^{-1}B_\text{multi}^\intercal P$, and $\mathbf{z}_\text{des}$ represents the target positions and velocities. Here $P$ is the solution to the continuous-time algebraic Riccati equation (CARE):
\begin{align*}
     A_\text{multi}^\intercal P+PA_\text{multi} - PB_\text{multi}R^{-1}B_\text{multi}^\intercal P + Q = 0.
\end{align*}
In addition, we employ APF method that introduces a potential field around each robot that creates repulsive forces to avoid collisions. The repulsive potential between robots $i$ and $j$ is given by:
\begin{align*}
    U_{\text{rep},ij}(z_i,z_j) = \left\{
    \begin{array}{cc}
       \frac{1}{2}k_\text{rep}\left(\frac{1}{d_{ij}}-\frac{1}{d_{\min}}\right)^2  &  \text{if} \quad d_{ij}< d_{\min}\\
       0 & \text{if} \quad d_{ij}\geq d_{\min}
    \end{array}\right.
\end{align*}
where $k_{\text{rep}}$ is a positive constant, $d_{ij}=\|p_i-p_j\|$ is the Euclidean distance between robots $i$ and $j$, and $d_{\min}$ is the minimum allowable distance between robots. The final control input for each robot, incorporating both LQR control and APF-based collision avoidance, is given by:
\begin{align*}
    u_i(t) = -K_i(z_i(t)-z_{\text{des},i}) - \nabla_{p_i}U_i.
\end{align*}
This ensures that the robots not only follow their intended paths but also avoid collisions by adjusting their trajectories dynamically in response to nearby robots.

\section{Experiments}
\subsection{Datasets and Experimental Setup}
\begin{table*}[h]
\centering
\caption{Cost $(\times 10^3)$ evaluation  on test dataset with 5 tasks in LA, 10 robots \& 505 episodic tasks (the lower the better)}
\label{tab:comp}
\begin{tabular}{cccccccc}
\toprule
\multirow{25}{*} & \multirow{2}{*}{Test distibution} & \multicolumn{3}{c}{Uniform distribution data} & \multicolumn{3}{c}{Gaussian distribution data} \\ \cmidrule(l){3-8}   
 &  & \texttt{MRTAgent} & BFO & FIFO & \texttt{MRTAgent} & BFO & FIFO \\   \cmidrule(l){2-8} 
 & \multirow{5}{*}{Similar}  & {\bf 13.49  $\pm$ 0.44} & 16.03  &  16.69 &  {\bf 18.23 $\pm$ 0.47}  & 18.60 &  19.08 \\
 &  & \multicolumn{1}{c}{\bf 14.85 $\pm$ 0.15} & \multicolumn{1}{c}{17.72} & \multicolumn{1}{c}{18.35} & \multicolumn{1}{c}{\bf 18.66 $\pm$ 0.06}  & \multicolumn{1}{c}{19.06} & \multicolumn{1}{c}{19.43} \\
 &  & \multicolumn{1}{c}{\bf 13.72 $\pm$ 0.16} & \multicolumn{1}{c}{15.75} & \multicolumn{1}{c}{16.49} & \multicolumn{1}{c}{\bf 19.10 $\pm$ 0.01}	& \multicolumn{1}{c}{19.77} & \multicolumn{1}{c}{20.41} \\
 &  & \multicolumn{1}{c}{\bf 14.83 $\pm$ 0.26} & \multicolumn{1}{c}{16.97} & \multicolumn{1}{c}{17.21} & \multicolumn{1}{c}{\bf 18.40  $\pm$ 0.01} & \multicolumn{1}{c}{19.83} & \multicolumn{1}{c}{20.67} \\
 &  & \multicolumn{1}{c}{\bf 14.47 $\pm$ 0.18} & \multicolumn{1}{c}{15.95} & \multicolumn{1}{c}{16.40} & \multicolumn{1}{c}{\bf 18.67  $\pm$ 0.09} & \multicolumn{1}{c}{19.04} & \multicolumn{1}{c}{19.73} \\
\cmidrule(l){2-8} 
 & \multirow{5}{*}{\begin{tabular}[c]{@{}c@{}} Totally\\  Different\end{tabular}} & \multicolumn{1}{c}{\bf 18.89 $\pm$ 0.12} & \multicolumn{1}{c}{19.06} & \multicolumn{1}{c}{19.43} & \multicolumn{1}{c}{\bf 15.84 $\pm$ 0.56} & \multicolumn{1}{c}{17.72} & \multicolumn{1}{c}{18.35} \\
 &  & \multicolumn{1}{c}{\bf 19.50 $\pm$ 0.67} & \multicolumn{1}{c}{20.15} & \multicolumn{1}{c}{20.56} & \multicolumn{1}{c}{\bf 14.84 $\pm$ 0.44} & \multicolumn{1}{c}{17.25} & \multicolumn{1}{c}{17.83} \\
 &  & \multicolumn{1}{c}{\bf 18.88 $\pm$ 0.13} & \multicolumn{1}{c}{ 19.04} & \multicolumn{1}{c}{19.73} & \multicolumn{1}{c}{\bf 13.98 $\pm$ 0.43}	 & \multicolumn{1}{c}{15.75} & \multicolumn{1}{c}{16.49} \\
 &  & \multicolumn{1}{c}{\bf 18.77 $\pm$ 0.24} & \multicolumn{1}{c}{19.58} & \multicolumn{1}{c}{19.65} & \multicolumn{1}{c}{\bf 14.37 $\pm$ 0.38} & \multicolumn{1}{c}{16.92} & \multicolumn{1}{c}{17.31} \\
 &  & \multicolumn{1}{c}{\bf 18.53 $\pm$  0.42} & \multicolumn{1}{c}{20.43} & \multicolumn{1}{c}{21.03} & \multicolumn{1}{c}{\bf 14.82 $\pm$ 0.38} & \multicolumn{1}{c}{15.95} & \multicolumn{1}{c}{16.40} \\   \bottomrule
 
\end{tabular}
\end{table*}

\noindent The \texttt{MRTAgent} framework has been validated within warehouse environment settings, operating under several key parameters to ensure efficient functioning. Due to the absence of publicly accessible real-world datasets for similar problem scenarios, synthetic data has been employed to evaluate our \texttt{MRTAgent} framework. Each episode spans \(\tau\) time units, and the synthetic datasets used for both training and evaluation are structured as square-shaped 2D continuous spaces ranging from \([0, 64] \times [0, 64]\), with task origins, destinations, and robot locations confined within this range.

\noindent In our experiments, task concentration varies to reflect specific periods of the day when the majority of tasks accumulate. Accordingly, datasets are generated using Gaussian distributions with varying means and standard deviations.The task's starting and ending coordinates, along with the task generation times, are provided to the planner through a limited-size LA. This \texttt{MRTAgent} framework is evaluated under two configurations: 
\begin{itemize}
    \item Normally distributed task arrival times
    \item Uniformly distributed task arrival times.
\end{itemize}

\noindent For each configuration, datasets containing ~500 tasks per episode are generated according to the corresponding distributions. For instance, in the case of normally distributed tasks, the task generation times adhere to $\mathcal{N}(600,50)$, where 600 represents the mean task generation time. For training, we considered a fleet of 10 robots, with the LA window length fixed at 5. The robots' charging threshold is set at 30\%, meaning robots with a SOC below 30\% must dock for recharging before resuming task execution. The steady charging rate is calibrated to be $16\times$ the discharging rate.

The planner and executor agents are implemented using the PyTorch library in Python 3.8, with an Adam optimizer~\cite{kingma2014adam}, a discount factor of 0.99, a lambda value of 0.95, a learning rate of 0.0003, an entropy coefficient of 0.001, a value function coefficient of 0.0002, and a batch size of 32. The policy networks for both the planner and executor are trained using the cross-entropy loss function, while the value networks are fine-tuned using the mean squared error loss metric.
\subsection{Baselines}
To the best of our knowledge, no existing work in the literature concurrently addresses multiple aspects of MRTA simultaneously. In light of the absence of established approaches, we propose two suitable baselines.

\noindent \textbf{Brute-force optimal (BFO)} : In this approach, all task-robot pairs (within the LA) undergo an exhaustive brute-force evaluation of time duration required for task execution by the robots, determined using standard Euclidean distance. The algorithm then selects the robot-action pair that minimizes this time duration. While brute-force optimal approach represents a locally optimal solution, the exhaustive evaluation significantly amplifies the run-time, posing practical challenges. Despite, this baseline is frequently adopted in the literature as a reference point for evaluating decoupled task allocation and navigation methodologies~\cite{mosteo2007comparative,agrawal2022dc}.


\noindent  \textbf{FIFO} : The FIFO baseline employs a dual-tiered decision framework. The initial allocation involves selecting the task that entered the LA queue first, aiming to reduce pending tasks within the queue and allocating the robot that can complete it earliest to minimize TTGT ~\cite{humbert2015comparative,hassin2021self}. Due to its simplicity, the FIFO approach requires the least execution time among all the considered approaches.

\begin{table}[h]
\centering
\caption{Cost $(\times 10^3)$ comparison on Gaussian dist. dataset with 5 tasks in LA, 10 robots \& 505 tasks/episode}
\label{tab:reward_comparison}
\begin{tabular}{@{}p{1.7cm}@{}p{1.4cm}@{}p{1.2cm}@{}p{1.7cm}@{}p{1.4cm}@{}p{1cm}@{}}
\hline
\multicolumn{3}{c}{\textbf{Avg TRTO}} & \multicolumn{3}{c}{\textbf{Avg TTGT}} \\ \midrule
\texttt{MRTAgent} & {BFO} & {FIFO} & \texttt{MRTAgent} & {BFO} & {FIFO} \\ \midrule
\textbf{8.44} & {8.48} & 8.63 & \textbf{11.38} & 11.68 & 11.93 \\
\textbf{8.03} & {8.59} & 8.93 & \textbf{11.09} & 11.18 & 11.48 \\
\textbf{7.59} & {8.62} & 9.11 & \textbf{10.79} & 11.20 & 11.56 \\
\textbf{7.70} & {8.08} & 8.30 & \textbf{10.90} & 10.96 & 11.43 \\
\textbf{8.20} & {8.49} & 9.22 & \textbf{10.70} & 10.81 & 11.25 \\ \bottomrule
\end{tabular}
\end{table}


\begin{table}[h]
    \centering
    \caption{Cost $(\times 10^3)$ evaluation on Gaussian dist. dataset with 5 tasks in LA, 30 robots \& 505 episodic tasks}
    \label{tab:largR}
    \begin{tabular}{cccc}
    \midrule
      &  \texttt{MRTAgent} & {\begin{tabular}[c]{@{}c@{}}BFO \end{tabular}} & FIFO \\ \midrule
        \multirow{5}{*}{Gaussian dist. data} & {\bf 8.41} & 8.64 & 8.81 \\ 
        & {\bf 8.93} & 9.27 & 9.63 \\ 
        & {\bf 8.19} & 8.34 & 9.05 \\ 
        & {\bf 7.49} & 7.63 & 8.53 \\ 
        & {\bf 8.19} & 8.43 & 9.61 \\
        \bottomrule
    \end{tabular}
\end{table}

\begin{table}[h]
    \centering
    \caption{Cost $(\times 10^3)$ evaluation on Gaussian dist. dataset with 5 tasks in LA, 25 robots \& 505 episodic tasks}
    \label{tab:largR_wt_retrain}
    \begin{tabular}{cccc}
    \midrule
      &  \texttt{MRTAgent} & {\begin{tabular}[c]{@{}c@{}}BFO \end{tabular}} & FIFO \\ \midrule
        \multirow{5}{*}{Gaussian dist. data} & {\bf 9.25} & 9.56 & 9.76 \\ 
        & {\bf 10.05} & 10.37 & 10.44 \\ 
        & {\bf 9.34} & 9.57 & 9.92 \\ 
        & {\bf 8.95} & 9.42 & 9.69 \\ 
        & {\bf 8.70} & 9.13 & 9.32 \\
        \bottomrule
    \end{tabular}
\end{table}

\begin{table}[h]
    \centering
    \caption{Cost $(\times 10^3)$ evaluation on Gaussian dist. dataset with 5 tasks in LA, 10 robots \& 2005 episodic tasks}
    \label{tab:largT}
    \begin{tabular}{cccc}
    \midrule
      &  \texttt{MRTAgent}  & {\begin{tabular}[c]{@{}c@{}}BFO \end{tabular}} & FIFO \\ \midrule
        \multirow{5}{*}{Gaussian dist. data} & {\bf 223.44} & 237.2 & 279.38 \\ 
        & {\bf 215.81} & 229.54 & 274.59 \\ 
        & {\bf 218.44} & 230.08 & 277.88\\ 
        & {\bf 217.31} & 232.45 & 276.56 \\ 
        & {\bf 214.67} & 227.45 & 274.31 \\
        \bottomrule
    \end{tabular}
\end{table}

\subsection{Training Details}
\begin{figure}
    \centering
    \includegraphics[width=1\columnwidth]{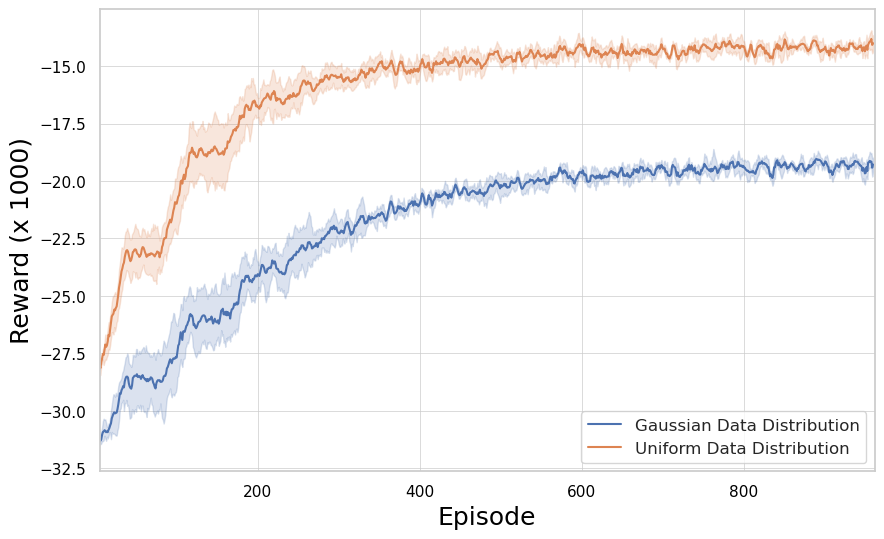}
    \caption{Learning curves of the MRTAgent}
    \label{fig:training profile}
\end{figure}
The \texttt{MRTAgent} is trained through a self-play approach. Initially, the planner undergoes training for 40 episodes, while the executor remains in evaluation mode. After every 40 episodes, the roles of the planner and executor are reversed: the planner is switched to evaluation mode, and the executor is trained. This cycle is repeated 24 times, leading to a total of 960 training episodes. Each episode consists of 505 tasks, with 10 robots in the environment and a LA length of 5. Both the Planner and Executor agents are trained using the PPO algorithm~\cite{schulman2017proximal}. The model is implemented using the PyTorch library in Python 3.8. Key hyperparameters include the Adam optimizer~\cite{kingma2014adam}, a discount factor of 0.99, a lambda value of 0.95, a learning rate of 0.0003, an entropy coefficient of 0.001, a value function coefficient of 0.0002, and a batch size of 32. The policy network for both agents is trained with the cross-entropy loss function, while the value network is optimized using mean squared error (MSE) loss. 

\subsection{Experiments and Results}
We now present a thorough comparative analysis of our proposed learning-based framework, \texttt{MRTAgent}, against baseline methods, namely the BFO approach and the FIFO strategy. Our experiments are designed to include scenarios that incorporate charging considerations, thereby addressing realistic operational challenges. The results consistently demonstrate the superiority of \texttt{MRTAgent} over the baseline methods.

Figure ~\ref{fig:training profile} presents the average training curve for the bi-level RL agents. This average is derived from four independent runs with different random seeds. The RL model is trained over 960 episodes, each comprising 505 tasks with 10 robots in the environment, and a look-ahead (LA) length of 5. The training process utilizes the PPO-based RL algorithm within the \texttt{PFRL} framework~\cite{JMLR:v22:20-376}. To simulate real-world task profiles, the training procedure employs task lists generated randomly from a Gaussian distribution, alongside uniformly sampled task lists. The reward plots indicate that \texttt{MRTAgent} achieves stable convergence towards an optimal policy. The improvements in rewards from their initial values suggest effective task selection and robot allocation, leading to reduced task waiting times within the LA window. It is worth noting that the RL agent is periodically trained with different random task lists, which prevents it from simply memorizing the performance on a specific task list. Instead, the agent learns to adapt to various scenarios, which explains the minor fluctuations in rewards across episodes as the agent refines its policy.


After the training phase, the model's performance is evaluated on different test datasets. For the model trained on task lists following a specific Gaussian distribution, the evaluation is performed on two distinct datasets: 
\begin{itemize}
    \item Instances that are similar to the training data with the same mean and variance specifically with $\mathcal{N}(600,50)$ distributed data,
    \item A dataset generated entirely from a uniform distribution specifically with $\mathcal{U}(0,1000)$ distributed data.
\end{itemize} 
This evaluation assesses \texttt{MRTAgent}'s ability to handle distributional shifts. Conversely, the model trained on uniformly generated task lists is evaluated on: 
\begin{itemize}
    \item A dataset sampled from the same uniform distribution specifically with $\mathcal{U}(0,1000)$ distributed data,
    \item A dataset generated from a Gaussian distribution pecifically with $\mathcal{N}(600,50)$ distributed data,
\end{itemize} 
offering insights into its performance under distributional shifts.


Table~\ref{tab:comp} provides a detailed comparison of the test results. The total number of tasks in each episode, the number of robots, and the LA queue length remain consistent with the training setup 505 tasks, 10 robots, and an LA length of 5. The results clearly demonstrate \texttt{MRTAgent}'s consistent outperformance over the brute-force optimal and FIFO-based methods across almost all test scenarios. The \texttt{MRTAgent} framework outperforms the baselines in all instances, and for a completely different dataset, it performs significantly better compared to the baselines.

To further analyze \texttt{MRTAgent}'s superiority over other baselines, we examine the individual reward components of all approaches, as shown in Table~\ref{tab:reward_comparison}. The total cost (reward) consists of two primary components: (a) TRTO, which aims to minimize robot travel distance, and (b) TTGT, which seeks to reduce task execution delays. As shown in Table~\ref{tab:reward_comparison}, \texttt{MRTAgent} achieves value, in the TRTO component lesser than the brute-force optimal approach signifying its ability to  minimize travel distance for the tasks in the look-ahead queue, as well as in the TTGT component because the brute-force optimal approach does not prioritize minimizing delays for tasks already in the look-ahead. On the other hand, the FIFO approach, despite assigning tasks sequentially, does not effectively reduce the TTGT component. This is because the task endpoints may be far from the starting locations of subsequent tasks in the look-ahead, potentially causing delays in task execution. As a result, the TRTO component is typically larger than in the brute-force.


\noindent\textbf{Variable number of robots}: To assess \texttt{MRTAgent}'s generalizability, we run experiments with varying numbers of robots during inference. Initially, the executor is trained with a fleet of 10 robots and 5 tasks in the LA, and the results are presented in Table~\ref{tab:comp}. We then retrain the executor for a fleet of 30 robots, while keeping the planner in evaluation mode, to observe \texttt{MRTAgent}'s performance and scalability. The results for the 30-robot scenarios are shown in Table~\ref{tab:largR}. As expected, performing the same tasks with more robots incurs lower costs, as the TRTO and TTGT components reduce significantly. Nevertheless, \texttt{MRTAgent} consistently outperforms the baseline methods in all scenarios.

The separation of Tasks (\textit{Planner}) and Robot (\textit{Executor}) agents in our framework enables scalability with varying task and robot counts. This setup also allows upgrading the \textit{Executor} without retraining the entire system. Scenarios like robot failures—which reduce available robot count—can be managed by extending robot availability time ($r_j$, a feature in \texttt{MRTAgent}) to a large value, excluding the failed robot from selection. For instance, \texttt{MRTAgent} trained with 30 robots (see Table~\ref{tab:largR}) is evaluated with only 25 available, without retraining in Table~\ref{tab:largR_wt_retrain}. This is achieved by assigning large values to $r_j$
for the extra 5 robots, effectively preventing task allocation and enabling our framework to adapt to different robot counts without retraining.

\noindent\textbf{Variable number of tasks}: To evaluate the \texttt{MRTAgent} framework's generalizability, the planner and executor, initially trained for a fleet of 5 robots and 505 tasks (see Table~\ref{tab:comp}), are tested on a larger number of tasks, specifically 2005 (see Table~\ref{tab:largT}). As observed, \texttt{MRTAgent} consistently outperforms the baselines across a variable number of tasks without requiring retraining.

In Tables~\ref{tab:comp}-\ref{tab:largT}, the different rows represent the use of various datasets for evaluation. Specifically, in Table~\ref{tab:comp}, the standard deviations reported for \texttt{MRTAgent} indicate slight fluctuations in performance when the model is evaluated using different random seeds. It is crucial to note that \texttt{MRTAgent} is trained with a periodically updated random task list, which prevents the agent from simply memorizing performance on a fixed set of tasks. Instead, this approach enables the agent to learn to adapt effectively to a wide range of scenarios. This is why there are small variations in the rewards across episodes, even as the agent gradually develops a consistent policy.

\noindent {\bf A note on the baselines}: While FIFO is known for its simplicity and computational efficiency; BFO, despite common intuition, is one of the strongest baselines which, given a look ahead (LA), evaluates all task-robot pairs and selects the one with earliest possible execution. In fact, BFO is an optimal task allocation and assignment approach given the current state of the LA received in an online fashion. The reason why \texttt{MRTAgent} is able to outperform it is due to the fact that \texttt{MRTAgent} exploits the underlying distribution defining task generation to plan for tasks to appear in future despite it having access to the same causal information as the BFO.


\section{Conclusion and Future Work}
This study introduces \texttt{MRTAgent}, a self-play motivated bi-level RL framework designed to enhance MRTA in modern warehousing environments. By optimizing operational costs and efficiency, \texttt{MRTAgent} addresses the increasing demands associated with online order fulfillment. The framework employs an optimal sequential task and robot selection process, coupled with a LQR based collision-free navigation algorithm. It effectively manages critical constraints such as robot dynamics, charging needs, and specific task requirements. Our approach demonstrates significant improvements over baseline methods across various test datasets. The validation of \texttt{MRTAgent} over datasets with distributional shifts and varying numbers of robots and tasks highlights its generalizability. 

While our \texttt{MRTAgent} algorithm presents a promising approach, it is essential to acknowledge certain limitations that warrant attention in future research endeavors:

\noindent \textbf{1. Single-Task Assumption}: The algorithm currently assumes that robots are engaged in one task at a time, potentially limiting its applicability in scenarios where multitasking is prevalent.
 
\noindent  \textbf{2. No Contingency for Sudden Breakdowns}: The model does not account for the sudden breakdown of a robot during operation. Once a task is allocated, our algorithm lacks the capability to reconsider or revert the decision in the event of an unforeseen robot malfunction.
 
\noindent \textbf{3. Homogeneous Robots}: We have made the simplifying assumption that all robots in the system are homogeneous, sharing identical values of characteristics such as velocity, acceleration, and load capacity. This assumption may not reflect the diversity present in real-world robot fleets.

\noindent  \textbf{4. Negligible Load/Unload Time Assumption}: The algorithm assumes negligible time for loading/unloading operations after a robot reaches the task origin/destination. In reality, this may not hold true, and accounting for realistic loading and unloading times is a consideration for future enhancements.

Future work will focus on enabling robots to handle multiple tasks simultaneously, incorporating load/unload times, integrating heterogeneous robots, and implementing learning-based navigation, all of which will enhance the algorithm's effectiveness and applicability in diverse, practical scenarios.



\bibliographystyle{ACM-Reference-Format} 
\bibliography{sample}


\end{document}